\documentclass[runningheads]{llncs}
\usepackage{graphicx}
\usepackage{float}
\usepackage{citesort}
\usepackage{algorithm}
\usepackage{algorithmic}
\usepackage{amsfonts}
\usepackage{url}

\DeclareMathAlphabet{\pazocal}{OMS}{zplm}{m}{n}
\newcommand{\Rs}{\pazocal{R}}
\newcommand{\Ws}{\pazocal{W}}
\newcommand{\Cs}{\pazocal{C}}

\begin{document}
%

\title{Marker based Thermal-Inertial Localization for Aerial Robots in Obscurant Filled Environments}
\titlerunning{Marker based Thermal-Inertial Localization}


%

%
\author{Shehryar Khattak \and Christos Papachristos \and Kostas Alexis}
%
%
\institute{Autonomous Robots Lab, University of Nevada, Reno, NV, USA
\url{http://www.autonomousrobotslab.com}}
\maketitle              
\begin{abstract}
For robotic inspection tasks in known environments fiducial markers provide a reliable and low-cost solution for robot localization. However, detection of such markers relies on the quality of RGB camera data, which degrades significantly in the presence of visual obscurants such as fog and smoke. The ability to navigate known environments in the presence of obscurants can be critical for inspection tasks especially, in the aftermath of a disaster. Addressing such a scenario, this work proposes a method for the design of fiducial markers to be used with thermal cameras for the pose estimation of aerial robots. Our low cost markers are designed to work in the long wave infrared spectrum, which is not affected by the presence of obscurants, and can be affixed to any object that has measurable temperature difference with respect to its surroundings. Furthermore, the estimated pose from the fiducial markers is fused with inertial measurements in an extended Kalman filter to remove high frequency noise and error present in the fiducial pose estimates. The proposed markers and the pose estimation method are experimentally evaluated in an obscurant filled environment using an aerial robot carrying a thermal camera.
\end{abstract}
%
%
\section{Introduction}\label{sec:intro}
Robotic inspection of infrastructure has seen an increasing interest over the past decade as it promises to mitigate risk to human life, minimize costs and reduce other disruptions frequently encountered during structural inspection tasks~\cite{roboticinspectionsurvey}. In particular, aerial robots, because of their advanced agility and flexibility have been applied to a variety of such inspection tasks~\cite{RHEM_ICRA_2017,bircher2018receding,montambault2010application,mascarich2018multi,papachristos2019autonomous,VSEP_ICRA_2018}. To navigate in known and unknown environments robots rely on reliable pose estimation information. Such information can be provided externally or estimated on-board using the data provided by sensors carried by the robot. External pose estimation is typically provided by Global Positioning System(GPS) in outdoor environments and by motion capture systems such as VICON or OptiTrack in indoor operations. However, GPS is limited to outdoor operations and suffers from multi-path inaccuracies in close vicinity to structures. Similarly, motion capture systems limit the work space of operation and require repetitive calibration for providing accurate pose estimates. Motion capture system can be also be very cost prohibitive. On-board a robot, pose can be estimated by utilizing sensors such as RGB camera systems, which due to their low weight and affordable cost are a popular choice. Using RGB camera images, visual odometry can be estimated reliably for local estimates, but tends to drift over time and relies on the quality of camera data. In addition, visual odometry estimates are local to robot's frame and do not provide any correspondence to a global map. Hybrid approaches, utilizing on-board sensing and the presence of previously known objects in the environment, such as fiducial markers, provide a low cost and reliable navigation solution to obtain global localization information. However, such methods still require quality camera data to localize the known markers~\cite{fidslam1,fidslam3}. As noted in~\cite{roboticinspectionsurvey,balta2016integrated}, not uncommon instances of inspection tasks are carried out in previously known environments in post-disaster conditions such as in the aftermath of a fire. These environments present a challenge for robot operations as they can be GPS-denied in nature and contain visual obscurants such as smoke. Although previously known environments provide the opportunity to have known markers placed in the environment for providing global positioning, yet in the presence of obscurants these disaster scenarios degrade RGB camera data significantly.
\par
\begin{figure}[h!]
\includegraphics[width=\textwidth]{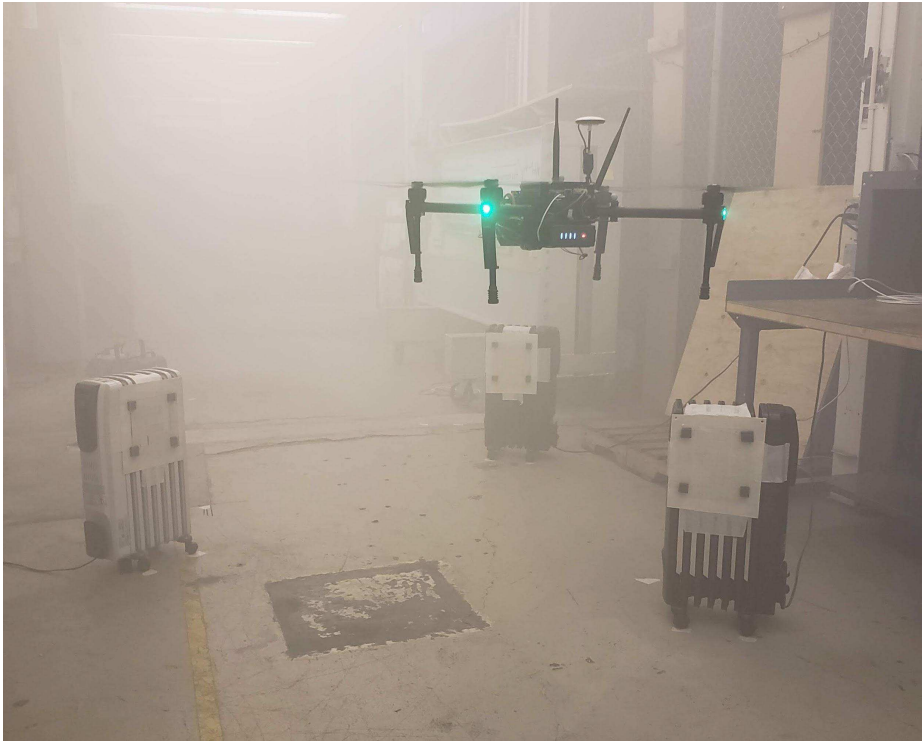}
\caption{Instance of an experiment during which an aerial robot equipped with a thermal camera navigates a fog filled room using thermal fiducial markers mounted on space heaters. Thermal fiducial markers are transparent in visible domain and only their corner foam mounting points can be seen in the image.}\label{fig:main}
\end{figure}

\noindent
Motivated by the challenging nature of such scenarios, in this paper we propose to extend the design and usage of fiducial markers into the Long Wave Infrared (LWIR) spectrum using thermal cameras for robot operation in obscurant filled environments. As opposed to the typical RGB cameras operating in the visible part of the electromagnetic spectrum, the selected thermal cameras operate in the LWIR part of spectrum and as a result, do not suffer from the same data degradation in the presence of certain obscurants, such as smoke and fog~\cite{ThermalPerception}. Similarly, utilizing the different optical properties of materials in the LWIR spectrum as compared to visible spectrum we extend the design of fiducial markers for LWIR spectrum usage while remaining unobtrusive in the visible spectrum. Furthermore, we augment marker based pose estimation by integrating inertial measurements for improved accuracy, higher update rate and accounting for frames with unreliable or missing marker detection. To verify the feasibility of the designed markers, detection of the designed markers with a thermal camera and inertial measurements fused pose estimation, an experimental study was conducted using an aerial robot in an obscurant filled environment. An instance of this study is shown in Figure~\ref{fig:main}.
\par\noindent
The remainder of the paper is structured as follows: Section~\ref{sec:method} details the design of thermal fiducial markers, the mapping algorithm and the pose estimation solution. Section~\ref{sec:experiments} presents the results of experimental study. Finally, conclusions are drawn in Section~\ref{sec:conclusion}.
\section{Proposed Method}\label{sec:method}
The proposed approach can be broadly divided into three sections, namely: i) Marker Design ii) Map Building and iii) Robot Pose Estimation, each of which are detailed below:

\subsection{Marker Design}
\begin{figure}[h!]
\includegraphics[width=\textwidth]{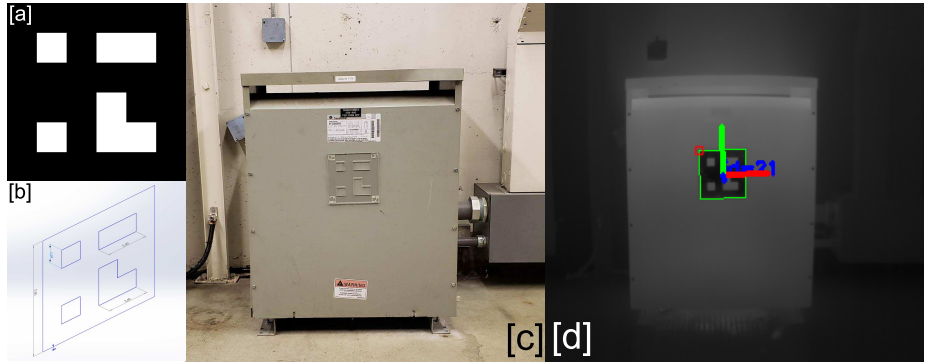}
\caption{The proposed marker design for thermal fiducial markers. [a] shows the original ArUco marker, [b] shows the generated CAD model for the marker, [c] shows the clear acrylic laser cut marker affixed to a transformer unit in a machine shop and [d] shows the detection of marker in the thermal image of the same scene.}\label{fig:marker}
\end{figure}

In the field of computer vision a wide variety of fiducial markers have been proposed for robust camera pose estimation~\cite{apriltag,artag,fiducialmarkers}. These markers are binary in nature and contain an internal coding of $ON/OFF$ bits represented by white and black colors respectively making the marker pattern unique and easy to identify. Although these markers can be in different shapes, square markers have the advantage of providing enough correspondences i.e. four corners from detection of a single marker to estimate the pose of the camera by solving the perspective-n-point problem. Similarly, knowing the dimension of one side of a square marker is enough for the accurate estimation of scale. For computer vision applications these markers are usually printed on a piece of white paper and affixed to a flat surface with the white paper providing the $ON$ bits and black printer ink providing $OFF$ bits. However this approach cannot be directly used in the presence of obscurants or with a thermal camera (responding to LWIR) without re-designing the fiducial marker. For the implementation of fiducial markers in the LWIR spectrum, we exploit the fact that thermal images respond and measure based on the emitted infrared radiation from objects of different temperatures and different material properties and encode this information on the the grayscale image space. Hence in our marker design, LWIR should be allowed to pass through at the locations of $ON$ bits of the marker and be blocked at the locations of $OFF$ bits of the marker. It is also important to maintain the flatness of the marker so it does not warp and the distance between the four corners is maintained. Similarly, the marker material should not be thermally conductive as to not heat up over time and become the same temperature as the surface to which it is affixed to making the marker pattern undetectable. Given the requirements of low thermal conductivity and the ability to be opaque to LWIR, \textit{acrylic} sheet was chosen as it is lightweight, low-cost, and transparent, making it a suitable material for such an application. Furthermore, it is a high tensile strength material enabling it to maintain its flatness and vaporizes during laser cutting process resulting in clean and precise cuts, similar to those obtained by printing commonly used fiducial patterns on paper, thereby allowing the use of those same patterns in the thermal domain. For our purposes we chose to use ArUco markers~\cite{aruco}, as they are square in shape, making the marker cutting process less complex. ArUco markers also allow us to create dictionaries of varying sizes and adjusting their intra-marker distances for better detection. A marker dictionary was created using the OpenCV implementation of the ArUco library and marker designs were exported as images. These images were first converted into vector graphics and then into CAD drawings for laser cutting. The process of marker creation as well its detection in thermal domain is shown in Figure~\ref{fig:marker}. It should be noted that our thermal fiducial marker design is very low cost and unobtrusive in visible spectrum. Our marker design can be accessed at (\url{https://tinyurl.com/LWIRMarkers})

\subsection{Map Building}
For robot localization in a known environment, a map is built by detecting markers attached to fixed objects in the environment. For map building we follow an approach is similar to~\cite{rosfiducialSlam}. The constructed map is represented in the world coordinate frame $(\Ws)$. This map is built in an incremental manner and adds new markers positions to the map only when they are mutually observed with markers previously defined in the map. When a set of markers is detected in a thermal image, first, we check which among the detected markers already exist in the map and then use them to estimate the camera pose in $\Ws$. Next, we estimate the pose of markers which do not exist in the map with respect to the camera coordinate frame $(\Cs)$. Finally, we calculate the pose of the new markers in $\Ws$ and add them to the map using their pose defined in $\Cs$ and the calculated camera pose defined in $\Ws$. We make two assumptions to build the map. First, the pose of at least one marker is known in $\Ws$ and this marker is used to start building the map. Second, at least two markers are visible in each thermal image as the environment is being mapped and new markers are being added. The first assumption allows for the building of a global map and estimate all marker positions in $\Ws$.
The second assumption only needs to be held true during the process of incrementally creating a map as adding new markers to the map requires calculation of the camera pose in $\Ws$. Once markers have been added to the map, the camera pose can be estimated by observing a single marker afterwards. Before adding a new marker to the map, we compute its re-projection error as a measure of its quality. This error is updated with each observation of the marker and is used as a weight when multiple known markers are observed in a single thermal image and used to determine camera pose in $\Ws$. Our marker map can be combined with other map representations as shown in~\ref{sec:experiments}, where we combine our marker map to a previously generated volumetric map. Such combinations can be used for other applications beyond the motivation of this paper e.g. offline path planning for inspection tasks using a volumetric map for collision avoidance and a marker map for viewpoint selection to improve localization~\cite{BABOOMS_ICRA_15}.
The steps of the map building process are summarized in Algorithm~\ref{alg:map}.
\begin{algorithm}[h!]
\caption{Map Building\label{alg:map}}
\begin{algorithmic}[1]
\STATE $\mathcal{MAP} \leftarrow$ Read Map
\STATE $\mathcal{I} \leftarrow$ Acquire New Thermal Image 
\STATE $\mathcal{D} \leftarrow \mathbf{DetectFidcuialMarkers}(\mathcal{I})$
\STATE $\mathcal{M}_{known},\mathcal{M}_{new} \leftarrow \mathbf{CheckKnownMarkers}(\mathcal{D},  \mathcal{MAP})$
\STATE $\mathcal{C}_{pose} \leftarrow \mathbf{CalculateCameraPose}(\mathcal{M}_{known})$
\STATE $\mathcal{E}_{reprojection} \leftarrow \mathbf{ComputeReprojectionError}(\mathcal{M}_{known})$
\STATE $\mathcal{E}_{previous} \leftarrow \mathbf{GetPreviousError}(\mathcal{M}_{known},\mathcal{MAP})$
\STATE $\mathcal{C}_{refined} \leftarrow \mathbf{RefineCameraPose}(\mathcal{M}_{known}, \mathcal{E}_{reprojection},\mathcal{E}_{previous})$
\STATE $\mathcal{MAP} \leftarrow \mathbf{UpdateError}(\mathcal{E}_{reprojection},\mathcal{E}_{previous})$
\FORALL{$\mathcal{M}_{new}$} 
    \STATE $\mathcal{M}_{pose} \leftarrow \mathbf{CalculateMarkerPose}(\mathcal{M}_{new},  {C}_{refined})$
    \STATE $\mathcal{M}_{error} \leftarrow \mathbf{ComputeReprojectionError}(\mathcal{M}_{pose})$
    \STATE $\mathcal{MAP} \leftarrow \mathbf{AddMarkertoMap}(\mathcal{M}_{pose}, \mathcal{M}_{error})$
\ENDFOR
\end{algorithmic}
\end{algorithm}
\subsection{Robot Pose Estimation}\label{sec:poseest}
Once we obtain the camera pose against a known map in $\Ws$, we can then calculate the pose of the robot by knowing the transformation between the camera coordinate frame $(\Cs)$ and the robot coordinate frame $(\Rs)$. As shown in~\cite{fidslam1,fidslam2,fidslam3} pose estimation from fiducial markers alone is sufficient to estimate the robot trajectory, however the accuracy of the estimated pose is subject to multiple factors such as the reliability of marker detection, the resolution of the camera, and the angle of marker observation. Also, as mentioned in~\cite{aruco}, the locations of the detected marker corners are prone to jitter if a corner is detected at lower resolution or from a large distance. This unreliable detection of corners causes the pose estimates of the robot to be noisy and less reliable which can then greatly impact onboard control. This is especially problematic for thermal cameras, as the commercially available and lightweight units operate at much lower resolution than their visible domain counterparts. Similarly, in some frames, marker detection can fail which causes jumps between pose estimates from fiducial. To make the robot's pose estimation robust, the estimated camera pose from the fiducial markers is fused with inertial measurements, obtained from the Inertial Measurement Unit (IMU) of the autopilot onboard the robot, using an extended Kalman filter (EKF). We use a formulation similar to~\cite{msf}, for our filter design where inertial measurements are used to propagate the state of the robot and pose estimates from the fiducial markers are used in the correction step of the filter. The state of our filter can be written as:
\begin{eqnarray}\label{state_eq}
\mathbf{x} = [\mathbf{p}~\mathbf{v}~\mathbf{q}~\mathbf{b}_f~\mathbf{b}_\omega]^T
\end{eqnarray}
 In our formulation, the robot coordinate frame $\Rs$ is aligned and centered with the IMU of the robot hence, $\mathbf{p}$ and $\mathbf{v}$ are the robot-centric position and velocity of the robot expressed in $\Rs$, $\mathbf{q}$ is the robot attitude represented as a map from $\Rs \rightarrow \Ws$, $\mathbf{b}_f$, $\mathbf{b}_f$ represents the additive accelerometer bias expressed in $\Rs$, $\mathbf{b}_\omega$ stands for the additive gyroscope bias expressed in $\Rs$. Proper IMU measurements i.e. the bias corrected, but noise affected accelerometer and gyroscope measurements are used to propagate the filter state. Camera pose measurements from the fiducial markers are then used in the correction step, where the differences between predicted and measured position $(\mathbf{p})$ and rotation $(\mathbf{q})$ states are used as an innovation term. This formulation helps in the reduction of noise and makes the pose estimation smooth, as well as enables the generation of robot pose estimates at the higher update rate of the IMU, making it suitable to be used for robot control tasks.
\section{Experimental Evaluation}\label{sec:experiments}
For the testing of the proposed markers and pose estimation method an experimental evaluation was conducted using an aerial robot carrying a thermal camera in an obscurant filled environment. The system and the experiment are detailed below:
\subsection{System Overview}
\begin{figure}[h!]
\includegraphics[width=\textwidth]{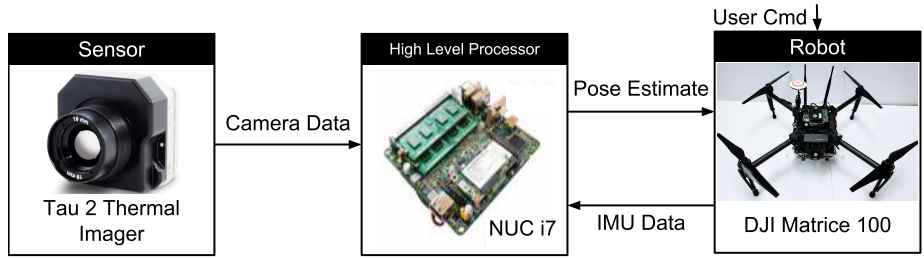}
\caption{High-level processor(Intel NUCi7) receives thermal images from Tau 2 camera and inertial measurements from the robot. The thermal images are processed for fiducial marker detection and initial pose estimation. The pose estimate is then integrated with inertial measurement using an Extended-Kalman Filter.}\label{fig:system}
\end{figure}

For the purpose of experimental studies a DJI Matrice 100 quadrotor was used. An Intel NUC i7 computer (NUC7i7BNH) was carried on-board the robot for performing all the high-level processing tasks including marker detection, map building and pose estimation. A FLIR Tau 2 thermal camera was mounted on the robot to provide thermal images of $640\times512$ resolution at $30$ frames per second. The intrinsic calibration parameters of the thermal camera were calculated using our custom designed thermal checker board pattern~\cite{icuas2018}. The robot autopilot IMU was used for providing inertial measurements. All algorithms were implemented as Robot Operating System (ROS) nodes and run in real-time fully on-board the robot. 
Figure~\ref{fig:system} provides the system overview of the robot system.
\subsection{Robotic Experiment}\label{sec:experiment}
\begin{figure}[h!]
\includegraphics[width=\textwidth]{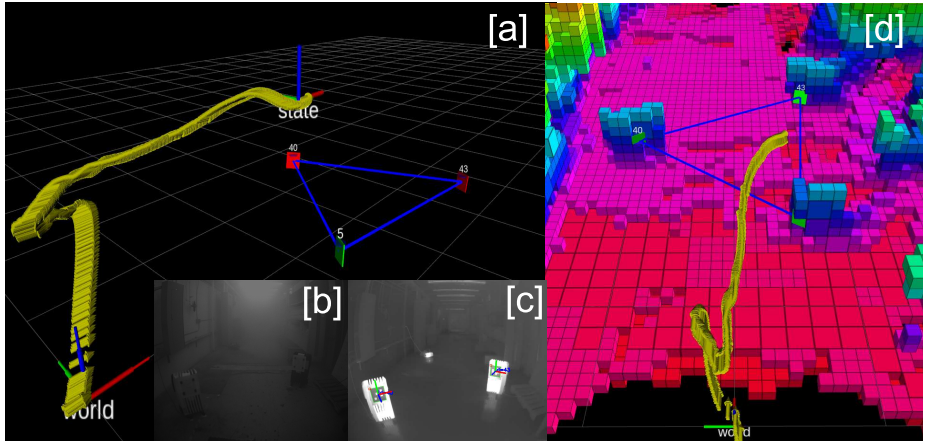}
\caption{A visualization of the robot trajectory during the experiment. [a] shows the robot trajectory in yellow, the fiducial markers are represented by green (unobserved) and red squares (observed). [b] and [c] show images from the visible and thermal cameras respectively at this instance during the robot trajectory. 
[d] shows the top view of the same trajectory with the fiducial marker map and volumetric map shown. On visual inspection it can be noted that the two marker maps are accurately aligned.}\label{fig:visualization}
\end{figure}

To evaluate the real-time on-board performance of the proposed approach, a test was conducted in an obscurant filled environment. In an industrial environment thermal markers are to be affixed to objects with a higher thermal signature with respect to their environment, an example is shown in Figure~\ref{fig:marker} where a thermal marker is affixed to a transformer unit. For the experiment, the designed thermal markers were affixed to space heaters in order to simulate an industrial setting. Our markers were cut from an $5$mm thick acrylic sheet. A fog generator was used to fill the testing environment with fog to serve as an obscurant. An instance of the experiment showing the transparent thermal markers mounted on a space heater and the obscurant filled environment is shown in Figure~\ref{fig:main}. A map for the fixed thermal markers was built and stored on-board the robot. To validate the accuracy of marker positions, a volumetric map of the environment was built separately in good visibility conditions using a pointcloud obtained from a visible light stereo camera along with localization provided by visual inertial solution ROVIO~\cite{rovio}. The two independently built maps align accurately as shown in Figure~\ref{fig:visualization}. A robot trajectory was executed with the same take-off and landing points, the mid-point of this trajectory is shown in Figure~\ref{fig:visualization}.
\par
\begin{figure}[h!]
\includegraphics[width=\textwidth]{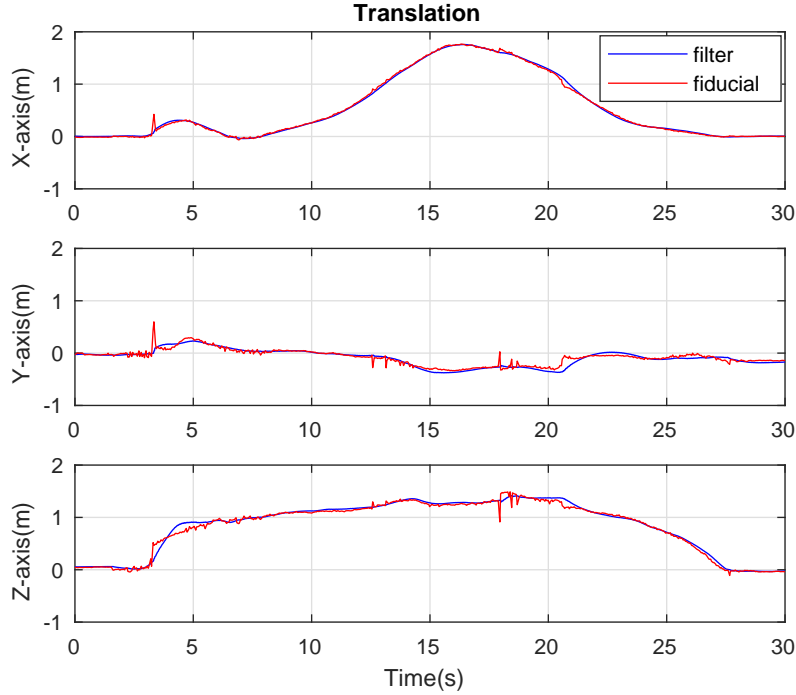}
\caption{The plot compares position estimation from the fiducial markers alone (red) and its integration with IMU measurement using an EKF (blue). The fiducial only position estimation is subject to jitters depending on the quality of observation. This is filtered out by fusion with inertial data.}\label{fig:plot}
\end{figure}

\noindent
To understand the improvement in pose estimation due to fusion of inertial measurements, the pose estimation from fiducial markers alone was compared to pose estimates generated by the implemented EKF described in Section~\ref{sec:poseest}. As shown in Figure~\ref{fig:plot} the pose estimation from fiducial markers alone is subject to jitter because of distorted marker observations and low resolution at long ranges. This jitter is filtered out after fusion with inertial measurements. A video of experimental results can be found at (\url{https://tinyurl.com/ObscuranctResults})



\section{Conclusions}\label{sec:conclusion}
In this paper we demostrated a method to design and extend fiducial markers from the visible spectrum to the LWIR spectrum to work with thermal cameras. Fiducial markers were designed and manufactured at very low cost and are minimally intrusive visually. The use of the designed markers was demonstrated by estimating the pose of an aerial robot in an obscurant filled environment.The pose estimation from the fiducial markers was made robust by fusion with inertial measurements using an EKF. Future work would consist of integrating the current solution with thermal-inertial navigation solutions for the robot to operate seamlessly between environments both containing and not containing markers. Similarly, in the future using visible and thermal markers in conjunction would be explored to make the solution more robust and generalizable to a variety of operating conditions and environments.

\bibliographystyle{ieeetr}
\bibliography{ISVC2018_FIDUCIAL}
\end{document}